\documentclass[conference,comsoc]{IEEEtran}
\usepackage{cite}

\ifCLASSINFOpdf
  \usepackage[pdftex]{graphicx}
\else
  \usepackage[dvips]{graphicx}
\fi
\usepackage[cmex10]{amsmath}
\usepackage{algorithmic}
\usepackage{array}
\ifCLASSOPTIONcompsoc
  \usepackage[caption=false,font=normalsize,labelfont=sf,textfont=sf]{subfig}
\else
  \usepackage[caption=false,font=footnotesize]{subfig}
\fi
\usepackage{dblfloatfix}
\usepackage{booktabs}
\usepackage{url}
\usepackage{float}
\usepackage{multirow}
\usepackage{enumerate}
\usepackage{amsthm}
\usepackage{threeparttable}
\usepackage{svg}
\usepackage{adjustbox}
\usepackage{booktabs,threeparttable}
\usepackage{array} 
\newcolumntype{C}[1]{>{\centering\arraybackslash}p{#1}} 

\newtheoremstyle{mystyle}
  {}
  {}
  {\itshape}
  {}
  {\bfseries}
  {.}
  { }
  {}
\theoremstyle{mystyle}

\newenvironment{talign*}
 {\csname align*\endcsname}
 {\endalign}
\usepackage{arydshln}
\usepackage[normalem]{ulem}
\usepackage{hyperref}
\usepackage{amsmath}
\usepackage{xurl}

\begin{document}

%
\title{Machine Learning for Cyber-Attack Identification from Traffic Flows}

\author{Yujing Zhou$^{1a}$, Marc L. Jacquet$^{1a}$, Robel Dawit$^{1a}$, Skyler Fabre$^{1a}$, Dev Sarawat$^{1a}$, Faheem Khan$^{1a}$, \\Madison Newell$^{1a}$, Yongxin Liu$^{1a}$, Dahai Liu$^{1a}$,Hongyun Chen$^{1a}$, Jian Wang$^{2b}$, Huihui Wang$^{3c}$\\
$^{1}$Embry-Riddle Aeronautical University, FL 32114 USA,\\
$^{2}$University of Tennessee at Martin, TN 38238 USA, $^{3}$Northeastern University, Arlington, VA 22209\\
    $^{a}$\{zhouy9, jacquetm,  dawitr, fabres, sarawatd, khanf10, newellm2\}@my.erau.edu, \\$^{a}${liuy11,liu89b,chenh4}@erau.edu
    $^{b}$jwang186@utm.edu, $^{c}$huih.wang@northeastern.edu\\
}
\IEEEtitleabstractindextext{%
\begin{abstract}

This paper presents our simulation of cyber-attacks and detection strategies on the traffic control system in Daytona Beach, FL. using Raspberry Pi virtual machines and the OPNSense firewall, along with traffic dynamics from SUMO and exploitation via the Metasploit framework. We try to answer the research questions: are we able to identify cyber attacks by only analyzing traffic flow patterns. In this research, the cyber attacks are focused particularly when lights are randomly turned all green or red at busy intersections by adversarial attackers. Despite challenges stemming from imbalanced data and overlapping traffic patterns, our best model shows 85\% accuracy when detecting intrusions purely using traffic flow statistics. Key indicators for successful detection included occupancy, jam length, and halting durations. All implementation details and source code are publicly available on GitHub at: https://github.com/U1overground/Cybersummer

\end{abstract}

}

\IEEEoverridecommandlockouts
\maketitle
\IEEEdisplaynontitleabstractindextext
\IEEEpeerreviewmaketitle

\section{Introduction}

As traffic control systems increasingly integrate automated and interconnected technologies, the risk of cyberattacks targeting both traffic flow and public safety has substantially increased threatening operational continuity and passenger safety. Traffic management systems, especially traffic lights, are prime targets for adversarial actors aiming to disrupt urban mobility and impose significant pressure on urban transportation flow \cite{feng2022cybersecurity}. By targeting traffic lights and similar control systems, attackers could easily disrupt emergency response times, delay public transportation schedules, and impact the daily commutes of thousands of citizens, leading to both economic losses and potential safety hazards \cite{ozarpa2021cyber}. Consequently, securing traffic management infrastructure has become crucial in modern urban planning, as any compromise can have cascading effects on the operational resilience, safety, and overall functionality of a city. 

The protection of urban traffic management systems against cyber threats requires a realistic model of targeted traffic systems, a.k.a. a digital twin \cite{jafari2023review}. This model would simulate both the physical infrastructure of the traffic flow and the cyber-physical systems that control and monitor it. By creating an integrated simulation environment, transportation planners and cybersecurity teams could test various scenarios, evaluate resilience under various cyber-attack conditions, and optimize response strategies. However, from the perspective of transportation cybersecurity, no integral simulation framework is readily available. Moreover, from the perspective of transportation practioners, raw data from network layers are not always available, making it harder for investigation and diagnosis against cyber attacks.

In this paper, we propose a detailed network model of the traffic system in Daytona Beach, FL. This model captures the interdependence between traffic lights due to proximity, simulating the cascading effects that arise when a single traffic light is compromised, such as delays and disruptions in nearby lights. Each traffic light is represented by a virtual machine connected via an internal network, enabling realistic interactions both between lights and with the main simulation system. This setup allows us to simulate hacking actions and assess their impact on data collection and traffic flow. To determine the operational status of each light, we apply neural networks and statistical machine learning algorithms to assess the likelihood of being compromised based solely on traffic data can provide a secondary layer of defense, enhancing the robustness of the detection system. The contribution of this paper is as follows:
\begin{itemize}
    \item We provide a joint simulation framework for both cyberattacks and the response to the traffic system.
    \item We investigate whether traffic data alone can be used to detect certain cyber attacks targeting traffic lights using both statistical and deep learning approaches.
    \item We found that the Maxim Halting Duration and Jam Length are two most critical metric for identifying cyber-attacks.
\end{itemize}


\section{Related Work}
\label{sectRW}

Transportation systems are increasingly targeted in cybersecurity due to vulnerabilities that threaten critical infrastructure and public safety. Interconnected traffic control systems are particularly at risk because their networked structure and integration with external devices and sensors offer multiple attack vectors \cite{pundir2022cyber}. The rising use of automated traffic management amplifies security challenges, especially in complex urban environments, making these systems attractive targets for maximum disruption \cite{feng2018vulnerability, zhou2022survey}. Thus, there is an urgent need for resilient cyber-defense frameworks tailored to transportation infrastructures, where disruptions can cause significant financial losses and safety risks \cite{ma2021smart}.

From a Cyber-Physical Systems (CPS) perspective, transportation-related cyber-attacks include Denial of Service (DoS) attacks flooding networks \cite{gu2007denial}, data manipulation altering sensor information \cite{chen2016analysis}, replay attacks resubmitting old valid data \cite{syverson1994taxonomy}, command injection manipulating system instructions \cite{usama2024command}, and man-in-the-middle (MitM) attacks intercepting communications between sensors and controllers \cite{conti2016survey}. These attacks exploit vulnerabilities inherent in CPS, highlighting the necessity for integrated cyber-physical security measures.

Detecting such cyber-attacks requires anomaly detection methods capable of analyzing both cyber and physical data. Machine learning techniques, including Random Forests, neural networks, and ensemble models, effectively identify complex anomalies in network traffic, sensor data, and command sequences \cite{bhuyan2013network, corea2024explainable}. Given the real-time operational demands of CPS, detection frameworks typically employ immediate analytics for prompt alerts \cite{liu2021zero}. Additionally, hybrid methods combining rule-based and deep learning approaches address challenges of high-dimensional data and rare attack occurrences \cite{liu2021class, pang2021deep}.

Emerging methods also indirectly detect cyber-attacks by analyzing traffic flow patterns, noting anomalies like unusual vehicle occupancy or queue lengths that may signal compromised systems \cite{jyothi2024data}. However, effectiveness varies depending on traffic complexity and attack subtlety \cite{bawaneh2019anomaly}.

While cyber-attack and traffic simulation frameworks individually assess vulnerabilities and traffic dynamics respectively, their integration remains limited. Cyber frameworks simulate threats like phishing and malware to evaluate security postures \cite{ficco2017simulation}, while traffic simulators analyze mobility impacts of disruptions \cite{hou2014integrated}. This paper addresses this gap by integrating these simulations, investigating distinctive traffic indicators at high-traffic intersections through a case study in Daytona Beach, Florida, to enhance security measures for modern traffic signaling systems.

\section{Methodology}
\label{sectMM}
\begin{figure}[h]
    \centering
    \includegraphics[width=\linewidth]{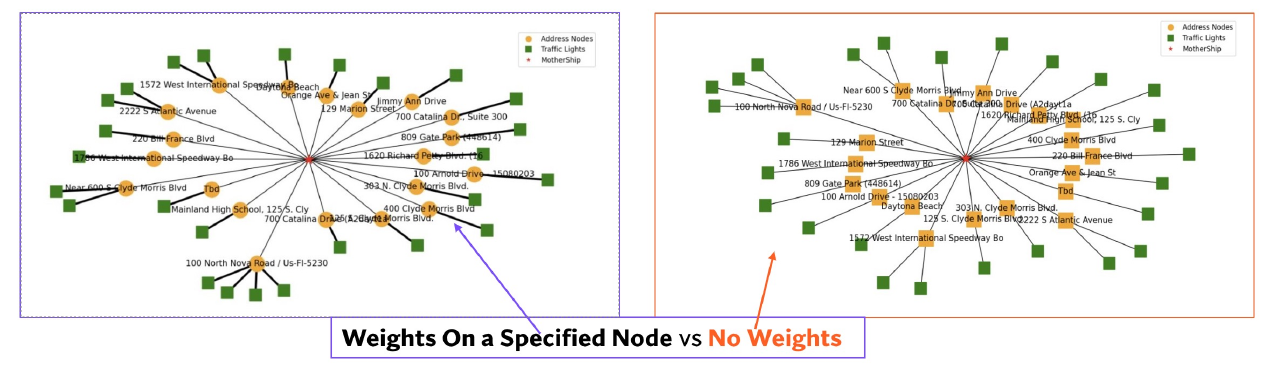}
    \caption{Abstract network topology of traffic controllers in Daytona Beach, FL}
    \label{figTrafficLightNetworks}
\end{figure}

We focus on simulating the cyber-attacks and defense strategies in the traffic control system of Daytona Beach, FL. We collected the base map tiles, location of traffic lights, locations of cellular towers where traffic light controllers are assumed use to connect to the Internet. We also retrieved the Annual Averaged Day Traffic Volumes of major roads for more precise modeling. After we finish the initial modeling, the abstract network topology is given in Figure~\ref{figTrafficLightNetworks}

The architecture of the simulation framework is shown in Figure~\ref{FigSimArch}. To make the simulation as realistic as possible, the following key technologies are employed:
\begin{figure}[h]
    \centering
    \includegraphics[width=\linewidth]{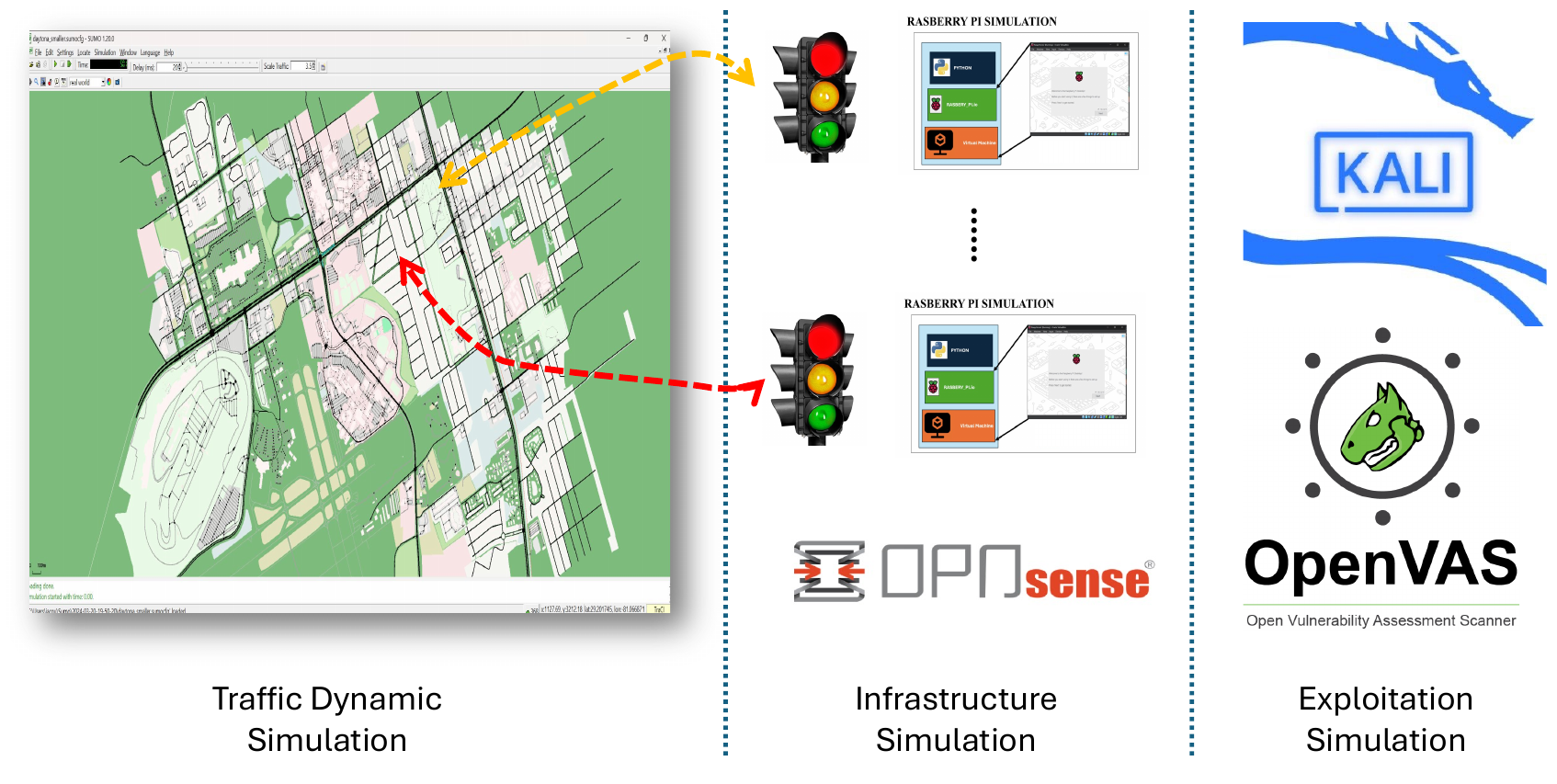}
    \caption{Architecture of the simulation framework}
    \label{FigSimArch}
\end{figure}

    \begin{itemize}
        \item Infrastructure Simulation: Raspberry PI Virtual Machines are used to seamlessly simulate the hardware and software environment of traffic control devices. We believe that the code in such scenario is practically deployable. Additionally, we use the \textit{OPNSense} (https://opnsense.org/), an open-source firewall, to create a dedicate and isolated virtual network to execute real cyber-attacks.
        \item Traffic Dynamic Simulation: Each Raspberry PI VM simulates one traffic light controller and connects remotely to \textit{SUMO} \footnote{~SUMO: https://eclipse.dev/sumo/} simulator via the \textit{TraCI} \footnote{~TraCI: https://sumo.dlr.de/docs/TraCI.html} interface. It is assumed that each traffic light controller only controls one intersection. We assume that each traffic light controller updates its phase pattern every 10 seconds.
        \item Exploitation Simulation: With the help of infrastructure simulation, we can directly use the \textit{Metasploit} \footnote{~Metasploit: https://www.metasploit.com/} framework and OpenVAS \footnote{~OpenVAS: https://www.openvas.org/} to simulate cyber-attack and defense events.
        
    \end{itemize}

\subsection{Statistical Methods Used for Classification}

Various machine learning algorithms were implemented and evaluated to predict the occurrence of hacks in the system based on sensor data. These algorithms included logistic regression, random forest, multilayer perceptron, k-nearest neighbors, and decision trees. The data were generated to simulate four distinct types of hack, together with a control simulation representing scenarios where no hack occurred. For the purpose of binary classification, all instances of hacked data were consolidated into a single class. This allowed for the training of binary classifiers capable of distinguishing between hacked and non-hacked states.

Traffic statistics data were collected over 10s intervals during the simulations. To ensure consistency and improve model performance, the numeric data generated from SUMO were normalized before being used to train the models. These models were selected for their ability to capture linear and polynomial relationships between inputs and outputs. Furthermore, their interpretability makes them particularly suitable for detailed analysis and understanding of the underlying patterns in the simulated data. All numerical values are scaled before being fitted to the logistic regression model, and all categorical data are one-hot encoded for training. 

\subsection{Convolutional Neural Network for Traffic Data Classification}
We employed a convolutional neural network (CNN) to classify traffic data and detect hacking incidents, testing input matrices of 9x23 (5s), 18x23 (10s), and 36x23 (20s) to identify the optimal temporal resolution.

\subsubsection{Data Preparation}
The dataset was preprocessed for CNN input as follows. First, irrelevant columns (target, begin, end, and ID) were removed. Specifically, begin and end columns were excluded due to overly high feature importance caused by their direct association with hacking intervals and events, introducing bias not representative of real-world conditions.

Remaining features were normalized using MinMaxScaler to a [0,1] range. Data was segmented into matrices of three configurations—9x23 (5s), 18x23 (10s), and 36x23 (20s)—with rows as sensor data snapshots and columns as 23 traffic features.

Labels were converted to binary form (1: normal traffic, 0: hacked traffic), aligned at intervals matching each matrix size. Finally, datasets were split into training and testing sets (80-20 split) and transformed into PyTorch tensors for CNN training.

\subsubsection{CNN Architecture}
\begin{figure}[h]
    \centering
    \includegraphics[width=\linewidth]{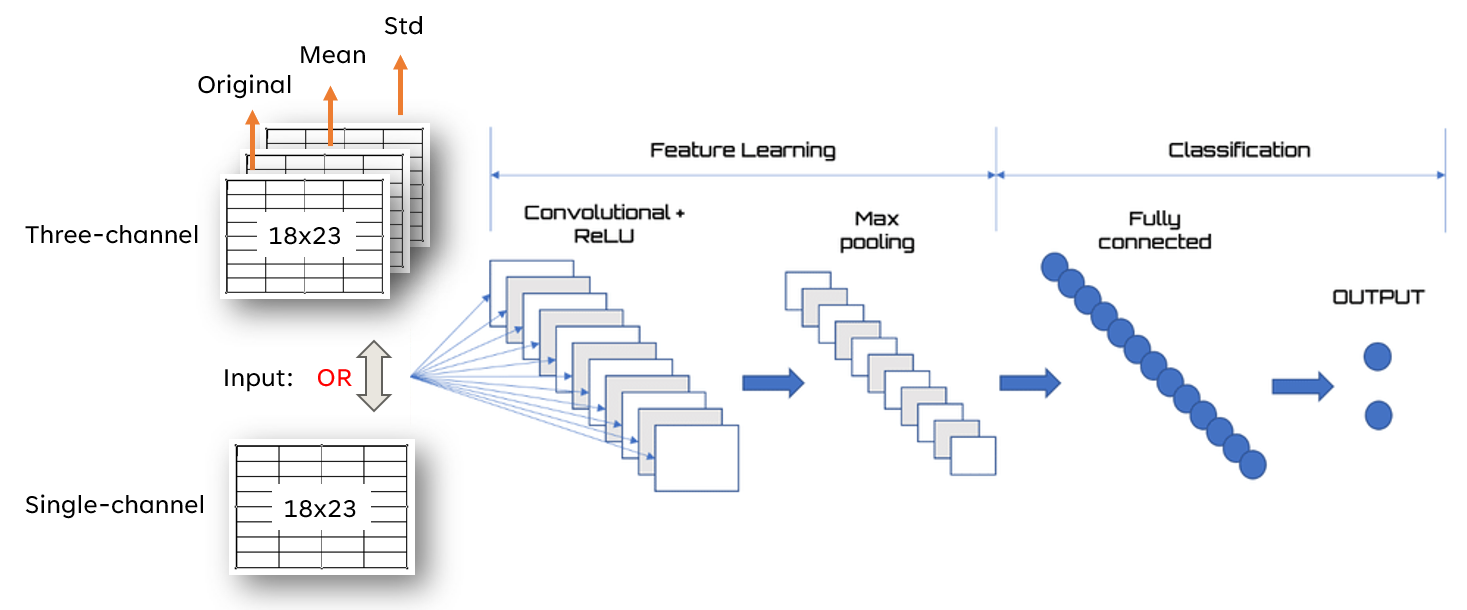}
    \caption{Architecture of the Convolutional Neural Network (CNN)}
    \label{FigCNNArch}
\end{figure}

The CNN architecture (CNN2D) used in this study is shown in Figure~\ref{FigCNNArch}. CNN2D processes two-dimensional inputs of sizes 9x23 (5s), 18x23 (10s), and 36x23 (20s), enabling analysis of how varying observation durations influence the detection of hacking events. The single-channel input is treated analogously to grayscale images, representing raw traffic data.

CNN2D consistently consists of two convolutional layers (64 and 128 filters, 3x3 kernels), each followed by a 2x2 max-pooling layer to reduce spatial dimensions. The output is then flattened, with dimensions calculated dynamically to accommodate different input sizes.

The fully connected layers include one hidden layer (64 neurons) and a single-neuron output layer with sigmoid activation for binary classification. This architecture provides flexibility and sufficient capacity to distinguish between normal and hacked traffic.

The CNN model was trained using Binary Cross-Entropy Loss and optimized with Adam (learning rate = 0.001) for 10 epochs with mini-batches of size 32. Data was shuffled each epoch for better generalization. We evaluated performance using three input sizes (9×23, 18×23, and 36×23), calculating accuracy, precision, recall, and F1-score, and generating confusion matrices to visualize classification performance.

Additionally, we tested a three-channel input configuration that combined the raw traffic matrix with corresponding mean and standard deviation matrices (computed from 10-second intervals of normal data). This provided statistical context, helping the CNN more effectively detect abnormal patterns. The first convolutional layer was adjusted to handle three-channel inputs, while the rest of the architecture (two convolutional layers with 64 and 128 filters, 2×2 max-pooling layers, and fully connected layers) remained unchanged. This enhancement allowed the model to process richer input representations across all tested input dimensions. However, such modification does not improve the network performance as expected.

\section{Results and Discussion}
\label{sectEED}
\begin{figure}
    \centering
    \includegraphics[width=0.9\linewidth]{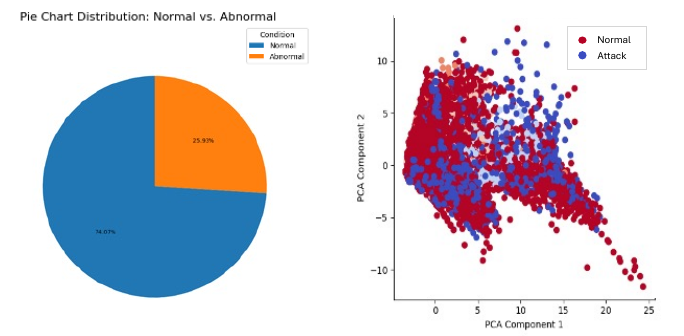}
    \caption{Data distribution of the simulated attack and normal scenarios. We used Principal Component Analysis (PCA) to project the original data to 2D space with 90\% of variance explained.}
    \label{figSimulationDataDistrib}
\end{figure}

We simulated the attack scenario that traffic lights are turned all green or all red randomly in the busiest intersection of Daytona Beach for one hour. Various traffic flow statistics are collected every 10 seconds, and these collected statistics form the initial dataset for an attack scenario. The distribution of the data w.r.t. different categories are shown in Figure~\ref{figSimulationDataDistrib}. 

As depicted, inferring cyber-attacks from traffic patterns could be challenging owing to two main factors: a) The dataset if highly imbalanced as the chance of observing a cyber-attack is rare and the collected dataset are highly imbalanced even after Synthetic Minority Over-sampling Technique (SMOTE) algorithm is applied to re-balance the dataset. b) As in Figure~\ref{figSimulationDataDistrib}, the data tuples representing attack scenarios are mixed almost uniformly with the normal scenarios in the 2D space generated by PCA, indicating the need of complicated feature engineering approaches. 

\subsection{Statistical Methods and Explanations}
To investigate the efficacy of traditional machine learning algorithms in detecting hacked traffic lights, we trained and evaluated six prominent models including Logistic Regression, Random Forest, SVM and etc. Each model was tuned using a feature set that captured various aspects of traffic flow, including average speed, queue length, and signal timing characteristics. Table I summarizes the key performance metrics for each model on the test dataset including accuracy, precision, recall, and F1-score, offering a clear comparison of their classification capabilities.

\begin{table}[h!]
    \centering
    \caption{Performance Metrics Comparison of Statistical Methods}
    \label{tab:crack_detection}
    \begin{tabular}{lcccc}
    \toprule
    \textbf{Models} & \textbf{Accuracy} & \(\textbf{Precision}\) & \(\textbf{Recall}\) & \(\textbf{F1-Score}\) \\
    \midrule
    SVM-SVC              & 0.704 & 0.68 & 0.70 & 0.62 \\
    Logistic Regression  & 0.684 & 0.60 & 0.68 & 0.58 \\
    Random Forest        & \textbf{0.757} & \textbf{0.73} & \textbf{0.75} & \textbf{0.74} \\
    KNN                  & 0.710 & 0.70 & 0.71 & 0.70 \\
    Decision Tree        & 0.720 & 0.71 & 0.72 & 0.71 \\
    MLP                  & 0.699 & 0.67 & 0.70 & 0.68 \\
    \bottomrule
    \end{tabular}
\end{table}

From Table I, we observe that Random Forest achieves the highest overall performance with an accuracy of 0.757, precision of 0.73, recall of 0.75, F1-score of 0.74. This suggests that ensemble-based methods capture complex relationships in the data more effectively than the other models tested. Decision Tree and KNN exhibit moderately strong metrics, indicating that both a hierarchical splitting strategy and instance-based classification capture meaningful patterns in detecting hacked traffic signals.

In contrast, Logistic Regression and MLP yield comparatively lower performance. Logistic Regression, for instance, may be constrained by its linear decision boundary, which could be a limiting factor if important relationships in the dataset are highly nonlinear. The MLP’s results suggest it may require further hyperparameter tuning or additional training data to fully exploit its nonlinear modeling capacity.

Notably, recall is a key metric in this application, since failing to detect a hacked signal (i.e., a false negative) can carry significant consequences for traffic safety. Random Forest’s strong recall (0.75) implies a relatively lower risk of missing compromised signals.

\begin{figure}[!ht]
    \centering
    \subfloat[]{%
        \includegraphics[width=0.9\columnwidth]{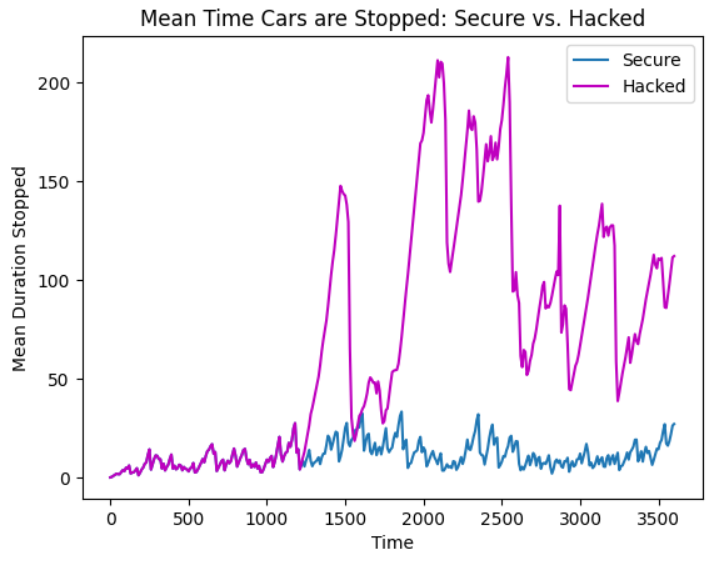}%
        \label{fig:max_halting_duration}
    }
    \\
    \subfloat[]{%
        \includegraphics[width=0.9\columnwidth]{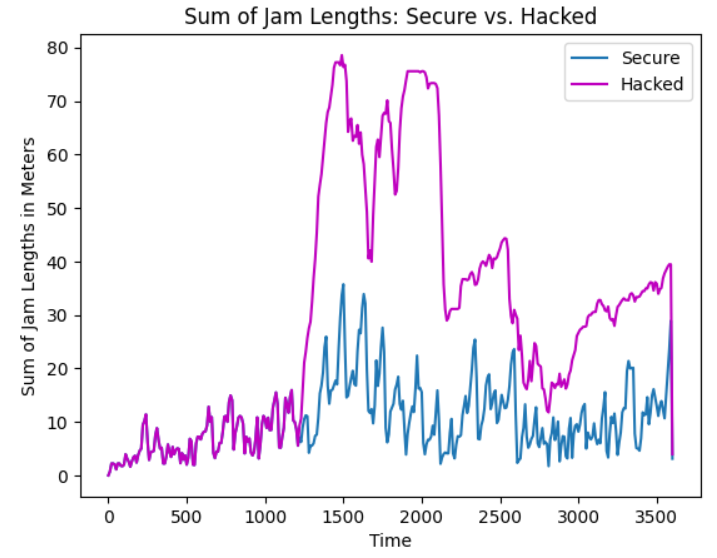}%
        \label{fig:jam_length_in_meters}
    }
    \caption{Two Most Important Features in response to intrusions in the Traffic Dataset}
    \label{fig:features}
\end{figure}

\begin{figure*}[!ht]
    \centering
    \subfloat[Logistic Regression]{%
        \includegraphics[width=0.48\textwidth]{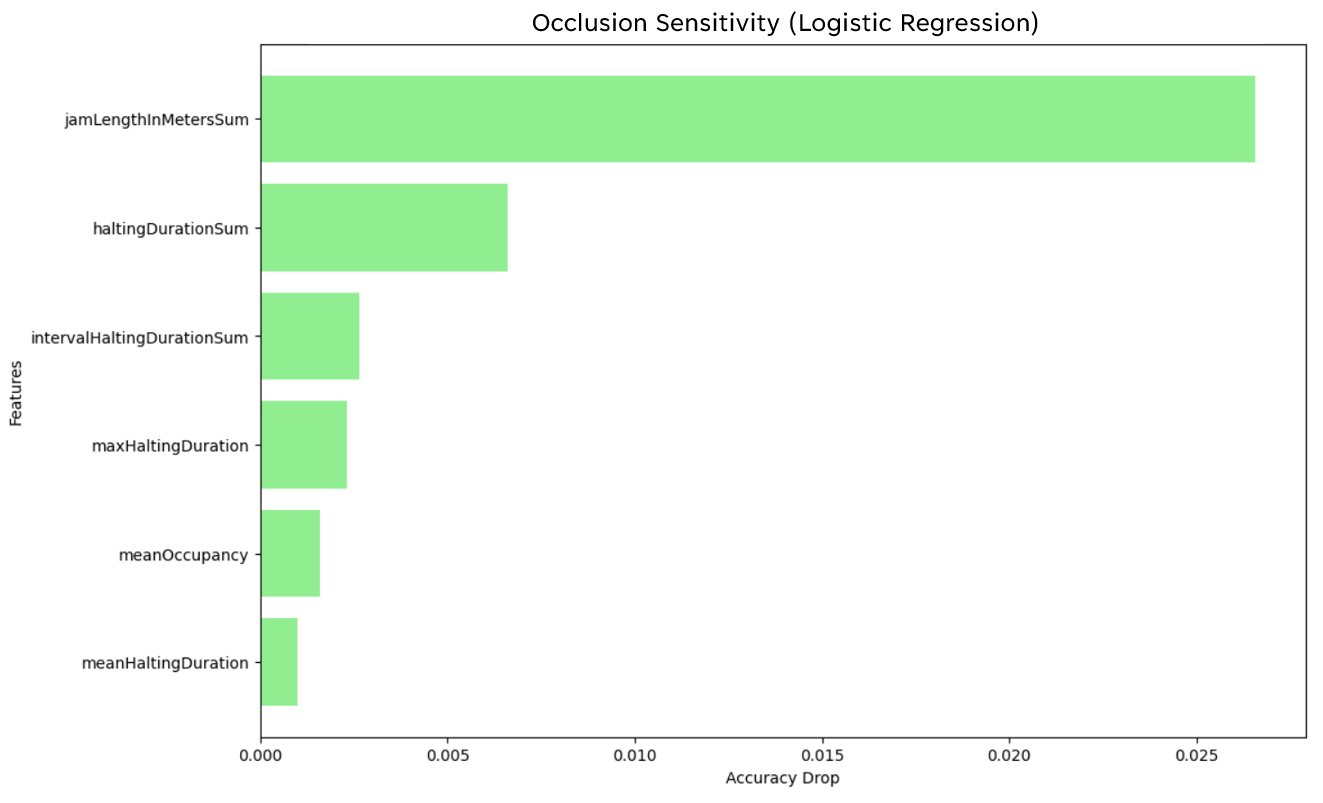}%
        \label{fig:logistic_regression}
    }
    \hfill
    \subfloat[Random Forest]{%
        \includegraphics[width=0.48\textwidth]{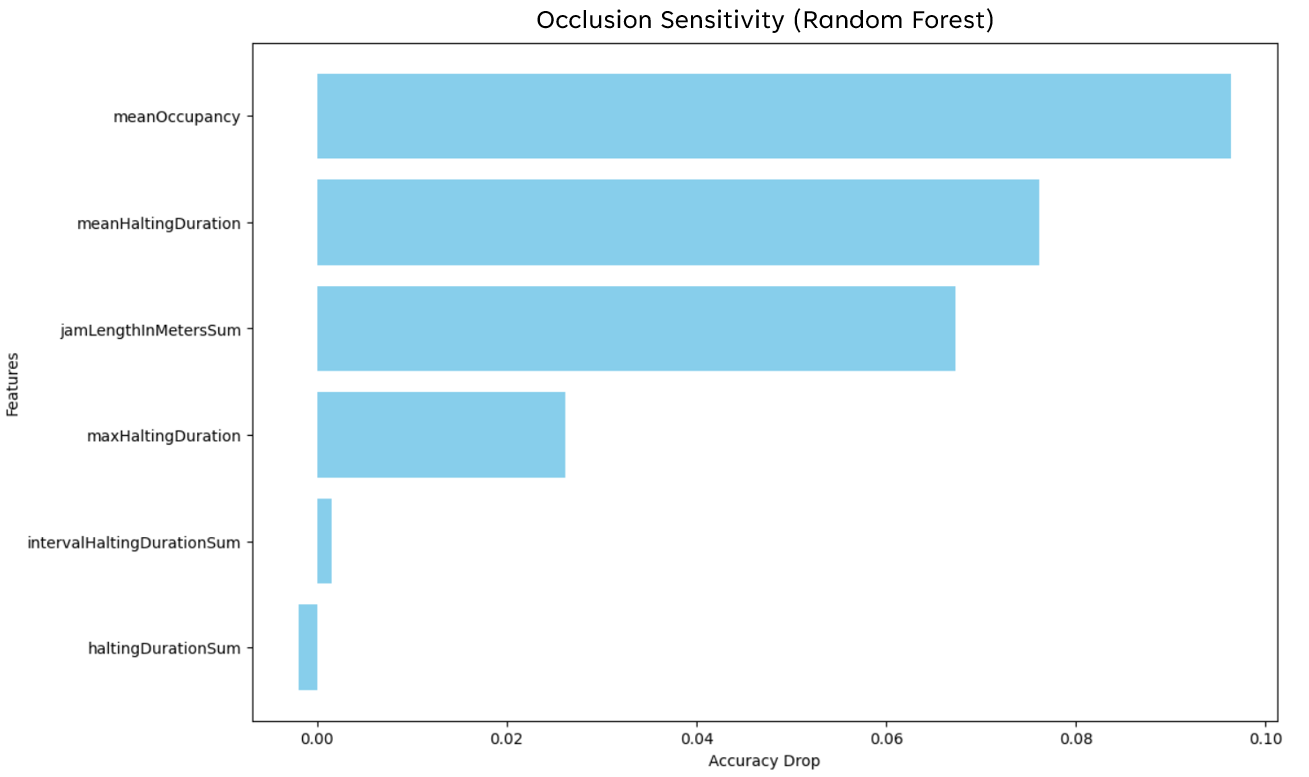}%
        \label{fig:random_forest}
    }
    \\
    \subfloat[KNN]{%
        \includegraphics[width=0.48\textwidth]{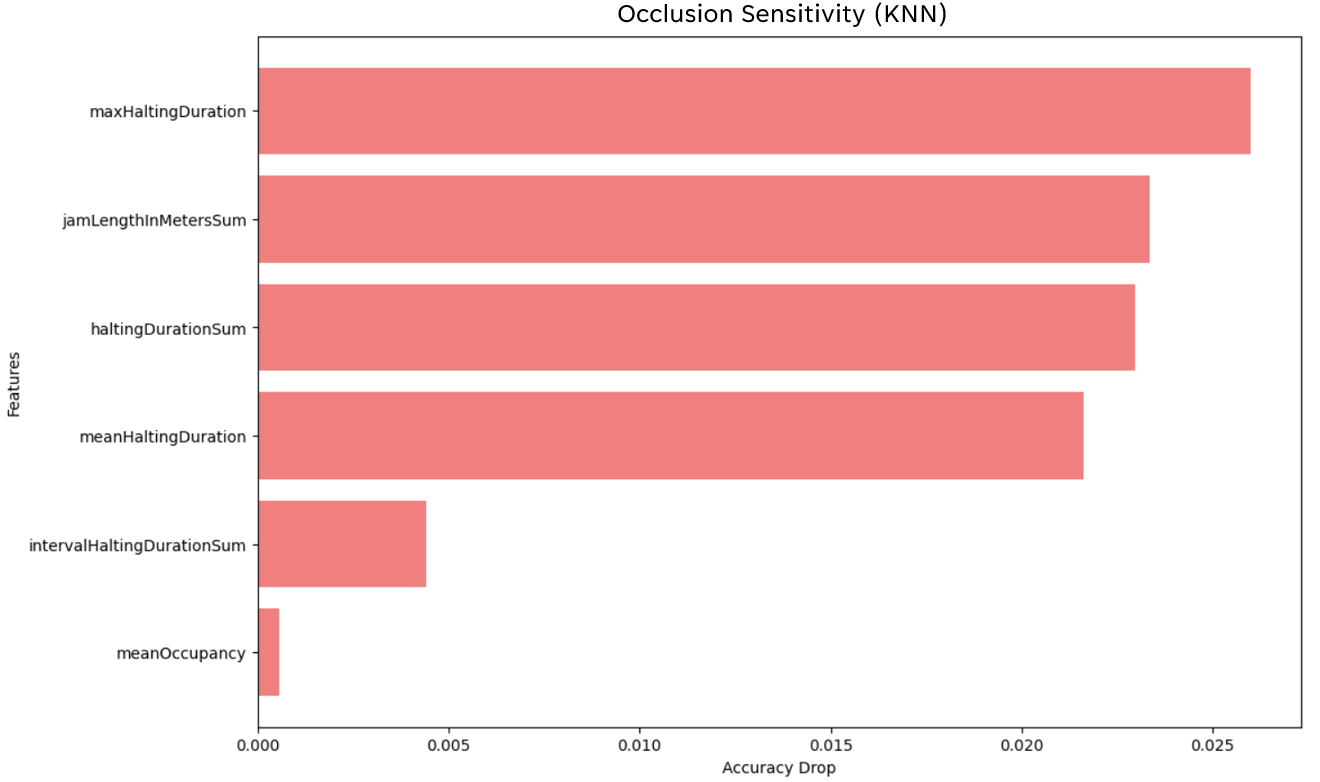}%
        \label{fig:knn}
    }
    \hfill
    \subfloat[MLP]{%
        \includegraphics[width=0.48\textwidth]{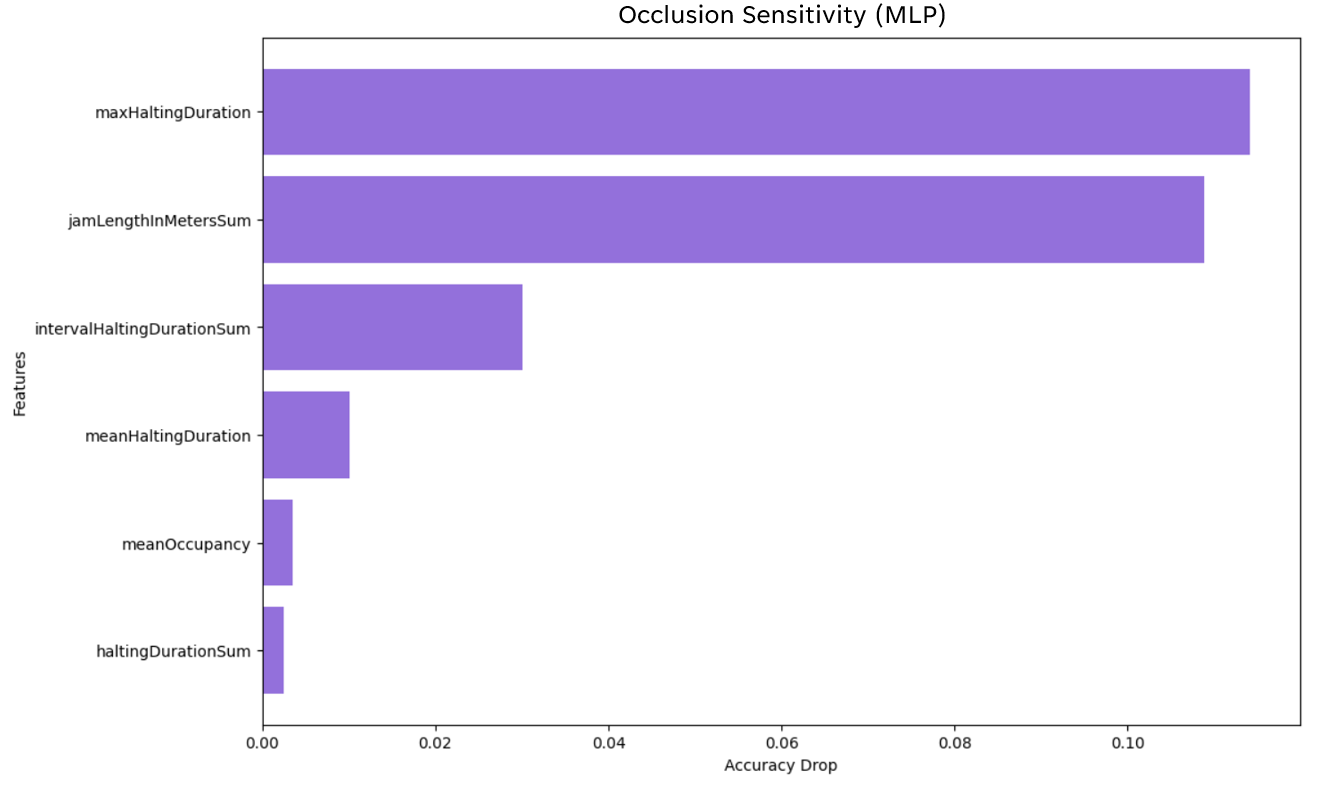}%
        \label{fig:mlp}
    }
    \caption{Occlusion Sensitivity Analysis for Different Models}
    \label{fig:occlusion_sensitivity}
\end{figure*}

 Across all models, congestion-related features (Figure~\ref{fig:features}), especially \emph{maxHaltingDuration} and \emph{jamLengthInMetersSum} tend to produce the largest accuracy drops, indicating that extreme standstill durations and significant traffic jams are strong indicators of hacked conditions. 

Figure~\ref{fig:occlusion_sensitivity} provides an occlusion sensitivity overview for each of the four models, illustrating how removing one feature at a time impacts classification accuracy. Logistic regression places the greatest emphasis on total jam length and overall halting duration, whereas random forest relies heavily on average occupancy and average halting time. KNN is most affected by the longest halting time and total jam length, while MLP similarly prioritizes maximum halting duration and cumulative halting intervals. 

Overall, any measure that captures unusual or prolonged standstills (\emph{maxHaltingDuration} and \emph{jamLengthInMetersSum}) emerges as a critical factor across all algorithms, suggesting that persistent congestion aligns closely with hacked behavior (Figure~\ref{fig:features}). Among these approaches, Random Forest shows the most balanced reliance on multiple signals and achieves the highest predictive accuracy, thus standing out as the most robust model for detecting hacked traffic signals.

\subsection{Convolution Neural Networks}
Although our statistical methods reaches an accuracy of around 0.75, the complexity of detecting hacked traffic signals suggest that a more sophisticated approach could offer additional improvements. CNNs are particularly well suited to learning complex patterns directly from large datasets, especially those with strong temporal or spatial components. So in this subsection, we compare its performance against the statistical methods discussed earlier.

\begin{figure*}[!t]
    \centering
    \includegraphics[width=2\columnwidth]{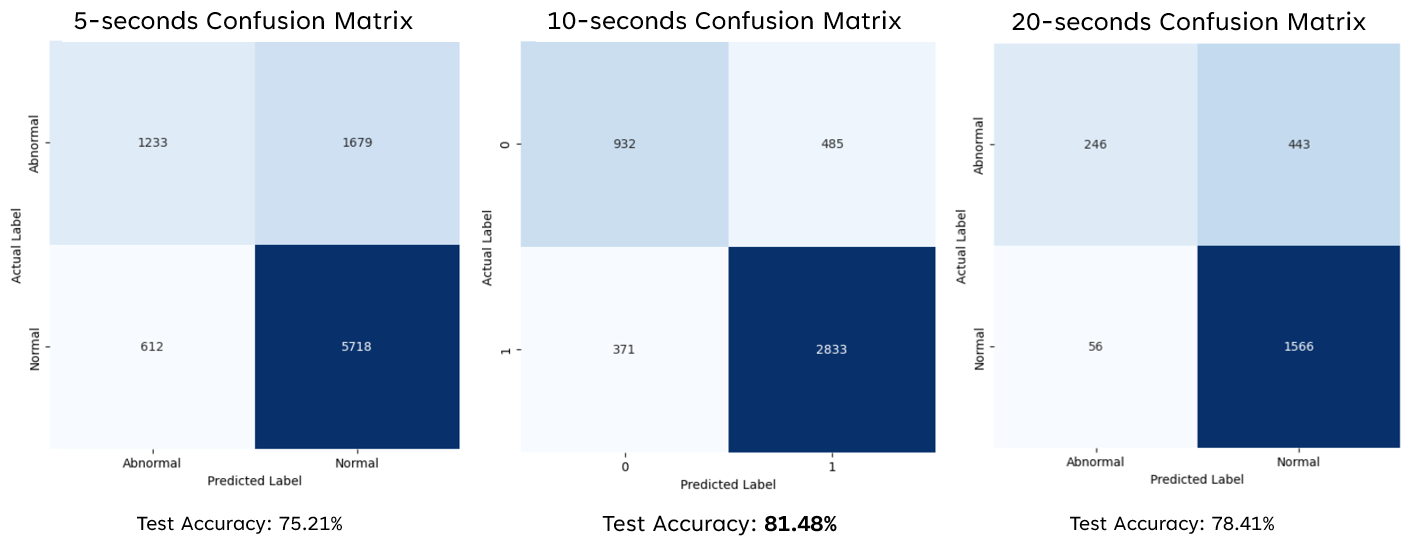}
    \caption{Confusion Matrix of CNN with Various Input Sizes}
    \label{fig:CNNconfusion}
\end{figure*}

As illustrated in Figure~\ref{fig:CNNconfusion}, using a 5-second window (9×23 input size) leads to a relatively high number of incorrect classifications, particularly where “abnormal” signals are labeled as “normal.” This suggests that such a short time frame may not capture enough context for reliably identifying hacked conditions. Increasing the window to 10 seconds (18×23) substantially reduces both false negatives and false positives, resulting in an improved accuracy of 81.48\%. Extending the window further to 20 seconds (36×23) moderates some misclassifications but lowers the accuracy slightly to 78.41\%, likely reflecting that longer time spans may introduce unnecessary complexity or dilute the immediate indicators of hacking activity. Therefore, a 10s window appears to keep the balance and provide sufficient detail to detect sudden anomalies without overwhelming the model.

In the previous subsection, we noted that Random Forest offered the strongest performance among the statistical models, reaching about 0.75 accuracy. By contrast, our CNN surpasses this mark when using a 10-second input window, achieving an accuracy of 0.81 (see Figure~\ref{fig:CNNconfusion}). While shorter (5-second) or longer (20-second) windows yield slightly lower results, this mid-range frame appears to capture enough temporal information to enhance detection of hacked signals. Overall, these findings underscore the CNN’s advantage in extracting richer patterns from the traffic data and improving upon the best statistical method.

\section{Conclusion}

In conclusion, our study indicates that Random Forest outperformed other statistical approaches across precision, recall, F1-score, and accuracy. However, the CNN achieved superior results due to its advanced capability in capturing complex traffic behaviors. Despite these promising outcomes, our detection method relies on indirect evidence of cyber-attacks from traffic anomalies rather than directly monitoring the traffic-light infrastructure. A robust future framework should combine anomaly detection with direct hardware or network port monitoring. Moving forward, we will enhance realism by expanding each traffic controller's connection to multiple cellular towers, improving network robustness. To address the overhead associated with virtual machines, we will containerize our simulation environment using Docker for greater computational efficiency. Lastly, recognizing the limitations of supervised statistical methods, we plan to explore deep learning—particularly unsupervised techniques—to effectively detect evolving and complex cyber-attacks in real-time scenarios.

\label{sectCC}

\section*{Acknowledgment}

This research was supported by the the USDOT Tier-1 University Transportation Center (UTC) Transportation Cybersecurity Center for Advanced Research and Education (CYBER-CARE) (Grant No. 69A3552348332).

\bibliographystyle{IEEEtran}
\bibliography{ref.bib}

\end{document}